\documentclass{article}

\usepackage{arxiv}

\usepackage[utf8]{inputenc} 
\usepackage[T1]{fontenc}    
\usepackage{hyperref}       
\usepackage{url}            
\usepackage{booktabs}       
\usepackage{amsfonts}       
\usepackage{nicefrac}       
\usepackage{microtype}      
\usepackage{graphicx}
\usepackage{enumitem}
\usepackage[numbers]{natbib}
\usepackage{doi}
\usepackage{lipsum}

\usepackage{multirow}

\title{Leveraging Large Language Model as Simulated Patients for Clinical Education}

\author{
    Yanzeng Li$^1$, Cheng Zeng$^{2,3}$, Jialun Zhong$^1$, Ruoyu Zhang$^1$, Minhao Zhang$^1$, Lei Zou$^{1}\thanks{Corresponding Author}$  \\ 
    $^1$Wangxuan Institute of Computer Technology, Peking University \\
    $^2$School of Computer Science, Wuhan University.
    $^3$CureFun Co.
}

\begin{document}
    \maketitle

    \begin{abstract}
        Simulated Patients (SPs) play a crucial role in clinical medical education by providing realistic scenarios for student practice.
        However, the high cost of training and hiring qualified SPs, along with the heavy workload and potential risks they face in consistently portraying actual patients, limit students' access to this type of clinical training.
        Consequently, the integration of computer program-based simulated patients has emerged as a valuable educational tool in recent years.
        With the rapid development of Large Language Models (LLMs), their exceptional capabilities in conversational artificial intelligence and role-playing have been demonstrated, making them a feasible option for implementing Virtual Simulated Patient (VSP).
        In this paper, we present an integrated model-agnostic framework called CureFun that harnesses the potential of LLMs in clinical medical education.
        This framework facilitates natural conversations between students and simulated patients, evaluates their dialogue, and provides suggestions to enhance students' clinical inquiry skills.
        Through comprehensive evaluations, our approach demonstrates more authentic and professional SP-scenario dialogue flows compared to other LLM-based chatbots, thus proving its proficiency in simulating patients.
        Additionally, leveraging CureFun's evaluation ability, we assess several medical LLMs and discuss the possibilities and limitations of using LLMs as virtual doctors from the perspective of their diagnostic abilities.
    \end{abstract}

    \section{Introduction}

    Clinical medical education plays a pivotal role in training aspiring healthcare professionals by equipping them with the necessary skills, knowledge, and providing practical experience to deliver high-quality patient care.
    One essential component of clinical education is Simulated Patients (SPs), who are individuals trained to portray specific medical conditions, symptoms, and behaviors in a standardized and consistent manner.
    By interacting with SPs, students can practice their clinical skills, communication, and decision-making in a controlled and realistic environment~\cite{pascucci2014integrating, sanko2013establishing, gaba2007future, mesquita2010developing, ziv2006simulation}.
    While SPs have been reliable and invaluable in clinical education, its universality and popularity often be limited due to high costs associated with training and hiring qualified individuals~\cite{hillier2020standardization, Felix2019TypesOS}.
    Meanwhile, SPs themselves face risks, including physical discomfort and psychological stress in consistently portraying actual patients~\cite{bokken2004performance, newlin2011effect}.
    As a result, the Virtual Simulated Patient (VSP)~\cite{mousavinasab2021intelligent, lopez2008tool, furlan2021natural} has emerged as an innovative approach to address these limitations.
    VSP systems leverage comprehensive technology, such as natural language processing and conversational artificial intelligence~\cite{rodrigues2022review, montenegro2019survey, chen2020artificial}, to simulate actual patient encounters and provide students with realistic clinical scenarios for conversation and diagnosis.
    One recently advancement is the rise of large language models (LLMs), powerful artificial intelligence systems capable of processing and generating human-like text and conversations~\cite{zhao2023survey, min2023recent}.
    LLMs, such as GPT~\cite{achiam2023gpt}, Llama~\cite{touvron2023llama}, PaLM~\cite{anil2023palm}, have demonstrated remarkable capabilities in natural language understanding and generation, leading to their application in various domains, including digital medicine applications~\cite{yang2022large, thirunavukarasu2023large, mesko2023imperative, kung2023performance, abd2023large}.
    Recently, researchers have begun to explore the potential of LLMs as SPs, e.g., some of contemporaneous works have explored utilizing LLMs like ChatGPT in simulation-based training~\cite{chen2023llm, benitez2024harnessing, holderried2024generative, sardesai2024utilizing}.

    Undoubtedly, the integration of \textit{LLMs as SP} holds great promise for clinical education.
    Benefiting from large-scale pre-training and aligning with human preferences, LLMs enable remarkable abilities such as following instructions, analyzing text content, and recalling existed information from the context~\cite{dong2022survey}.
    These fundamental capabilities are essential for a qualified SP\@.
    However, the existing approaches, including LLM-based methods, face certain challenges~\cite{abd2023large}.
    One notable issue is hallucinations, which can result in the generation of fictional information and factual errors, thereby reducing the realism of clinical training.
    While existing LLMs primarily aim to align with the perspective of healthcare advisors to provide helpful responses in addressing users' concerns~\cite{shen2023large, han2024towards}, they often struggle to accurately portray an actual patient, leading to problems such as role flipping in SP conversations.
    Additionally, there are several general challenges in conversation AI, such as instruction leakage, flake replies, infinite repetition, and toxic responses~\cite{li2023camel, kasneci2023chatgpt}.
    These issues need to be addressed to enhance the performance and reliability of VSP. Another challenging aspect in the utilization of SPs for clinical education is the assessment of students' medical dialogues.
    In traditional clinical examinations involving SPs, the student-patient encounter is typically evaluated by the SPs and supervisors by a checklist, which assesses various aspects of the student's performance~\cite{ladyshewsky1999simulated, smith2014simulated}.
    Traditional manual or rule-based grading methods face a series of issues, including raters' imperfect subjective judgments, the inability to dynamically and comprehensively evaluate the whole conversation, and difficulty in scaling up to large-scale assessments~\cite{kaldaras2022validation, wang2008automated, messer2023automated}.

    \begin{figure}[ht]
        \centering{\includegraphics[width=.9\linewidth]{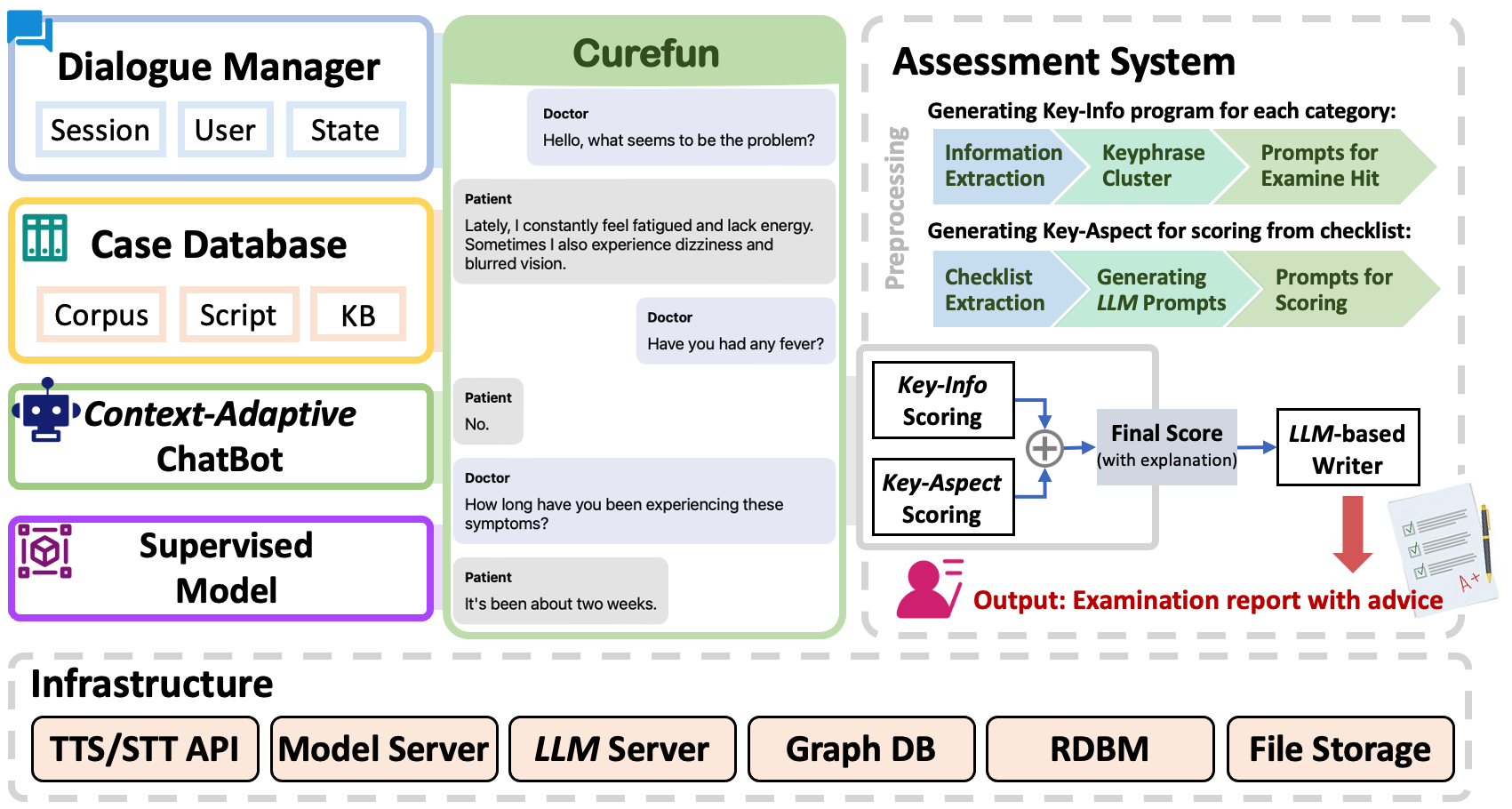}}
        \caption{The overview diagram of this study. Curefun integrates LLMs to simulate patient roles, enhancing the dialogue flow via structured graph memory, and providing automatic assessment for student-patient conversations.}
        \label{fig:overall}
    \end{figure}

    In this study, we present an integrated model-agnostic framework called CureFun that utilizes the comprehensive abilities of LLMs. Specifically, in the process of preparing the SP template, implementing student-patient conversation, and evaluating SP dialogue history, we utilize instructions such as Prompt and Chain-of-Thought to control the LLM behaviors according to our predetermined procedure.
    This allows us to construct a structured SP case graph, enhancing and controlling the dialogue flow in the form of retrieval-augmented generation~(RAG)~\cite{gao2023retrieval, lewis2020retrieval}.
    Finally, through the generated multi-granularity assessment items, supervised by multiple LLMs and their voting, we can grade an appropriate score to the student, and provide suggestions.
    In summary, this framework aims to address existing issues and develop a productive VSP application for clinical education.
    Through comprehensive evaluations, we demonstrate that our approach enables more authentic and professional dialogue flows in SP scenarios compared to other LLM-based chatbots, and shields most of the aforementioned flaws.
    Furthermore, we enhance CureFun's assessment capability by automatically converting traditional SP evaluation checklists into LLM-executable programs.
    We then ensemble multiple LLMs to collaborate and provide a comprehensive and reliable assessment of students' medical dialogues.
    This enables large-scale and efficient SP-involved assessment in clinical education.
    Moreover, leveraging CureFun's evaluation capability, we assess several LLMs and discuss the possibilities and limitations of using LLMs as Virtual Doctors (VDs)~\cite{thirunavukarasu2023large}.
    In conclusion, our exploration highlights the potential of LLMs as VSPs for more efficient clinical education.
    And, it provides in-depth insights into the development of medical LLMs for intelligent diagnosis and treatment.

    \section{Result}

    In this section, we present the results of our comprehensive evaluation of the proposed framework, CureFun, in the context of acting SPs, automatic assessment, and evaluating LLMs as VDs in our SP system.

    \subsection{The Preparation of SP Cases}

    We have meticulously selected 8 cases in the Chinese language, with each case representing an individual SP script.
    These cases cover a range of diseases, including gastric disorders, diabetes, chronic obstructive pulmonary disease (COPD), COVID, pneumonia, and bronchiectasis.
    Moreover, these cases span multiple medical specialties such as pulmonology, endocrinology, and gastroenterology.
    The dataset used in this study was obtained from Wuhan Talent Information Technology Co., Ltd.
    After performing data cleaning, preprocessing, and manual data selection, we identified these eight cases as representative and comprehensive high-quality data for our subsequent experiments.
    An example of one of our SP cases is illustrated in Figure~\ref{fig:case-example}.

    \subsection{Generation Quality}

    We conducted evaluation experiments on 8 SP cases.
    We employed experts who are proficient in SP education assessment.
    They engaged in conversations with six mature LLM chatbots, with all SPs being given the same role-playing prompt to start with.
    During the interaction with the SPs, the experts asking questions according to the context and necessity of diagnose, to ensure a smooth conversation.
    The conversations were terminated either when the maximum number of rounds (N=20) was reached or when the expert deemed it appropriate to end the dialogue.
    During the evaluation, the experts were unaware of which model was behind the tested VSP. In this study, we uses open-source LLMs with large parameter-scale and well-established commercial models as backbone models, owing to their advanced capacity to comprehend instructions and participate in dialogue~\cite{kaplan2020scaling, duan2023botchat}.
    Specifically, we include the following LLMs in our study: GPT-3.5-turbo, PaLM~\cite{anil2023palm}, ERNIE-4~\cite{sun2021ernie}, Mixtral-8x7B~\cite{jiang2024mixtral}, and Qwen-72B~\cite{bai2023qwen}.
    All of the included LLMs are capable of conversion in Chinese language, and chat versions of open-source LLMs are specially utilized in our investigation.

\begin{figure}[ht]
        \centering{\includegraphics[width=.65\linewidth]{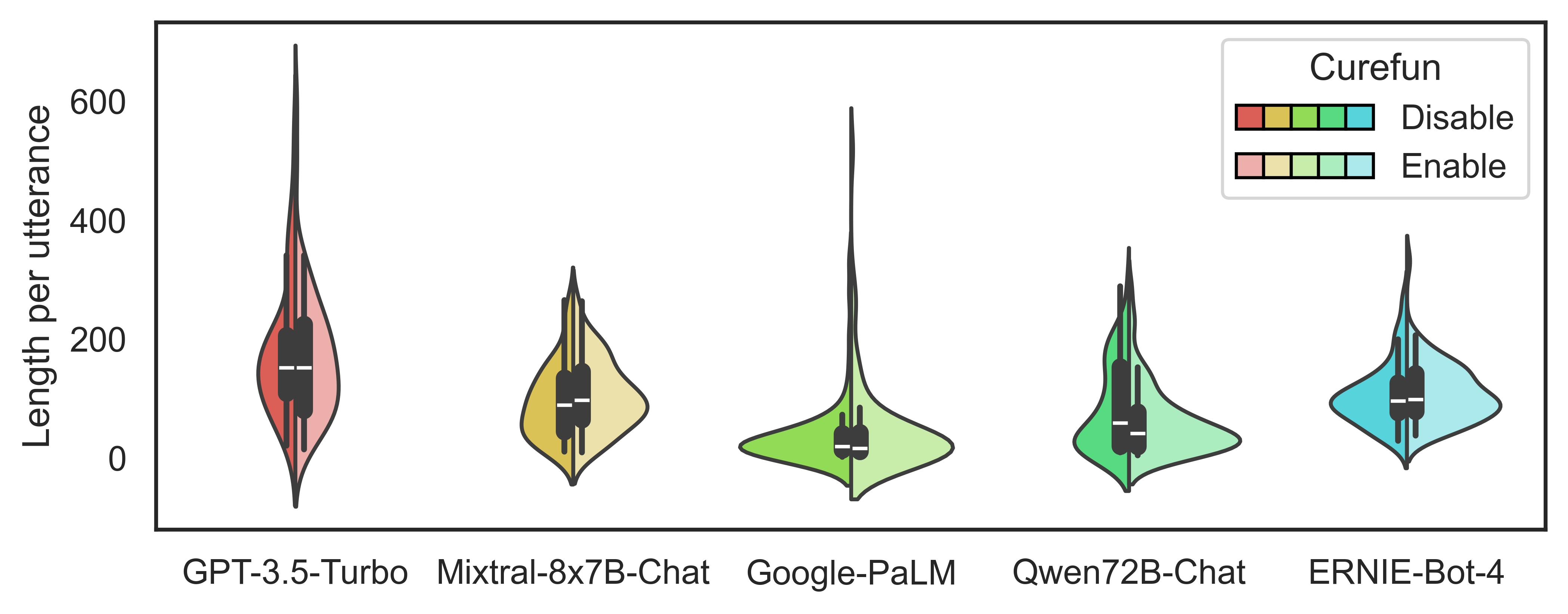}}
        \caption{Violin plot of the token length distribution of responses from different underlying LLMs when acting as SP and answering inquiries from doctors. The statistics on the left side of each violin represent the vanilla models' generation, while the right sides represent the responses generated with our proposed framework.}
        \label{fig:SP-length}
    \end{figure}

    In the LLM as SP experiment, we observed that the different LLMs exhibited varying characteristics based on the enlighten of the same prompt.
    Figure~\ref{fig:SP-length} presents the statistics of the response lengths generated by each LLM when playing the role of the patient and responding to the doctor's inquiries.
    It can be observed that PaLM often provides concise replies, whereas GPT-3.5-turbo tends to respond with more detailed content.
    After incorporating we proposed framework, there was no significant change in the distribution of response lengths, indicating that our framework would not visibly impact the inherent characteristics and personalities of the underlying LLMs.

\begin{figure}[ht]
        \centering{\includegraphics[width=.75\linewidth]{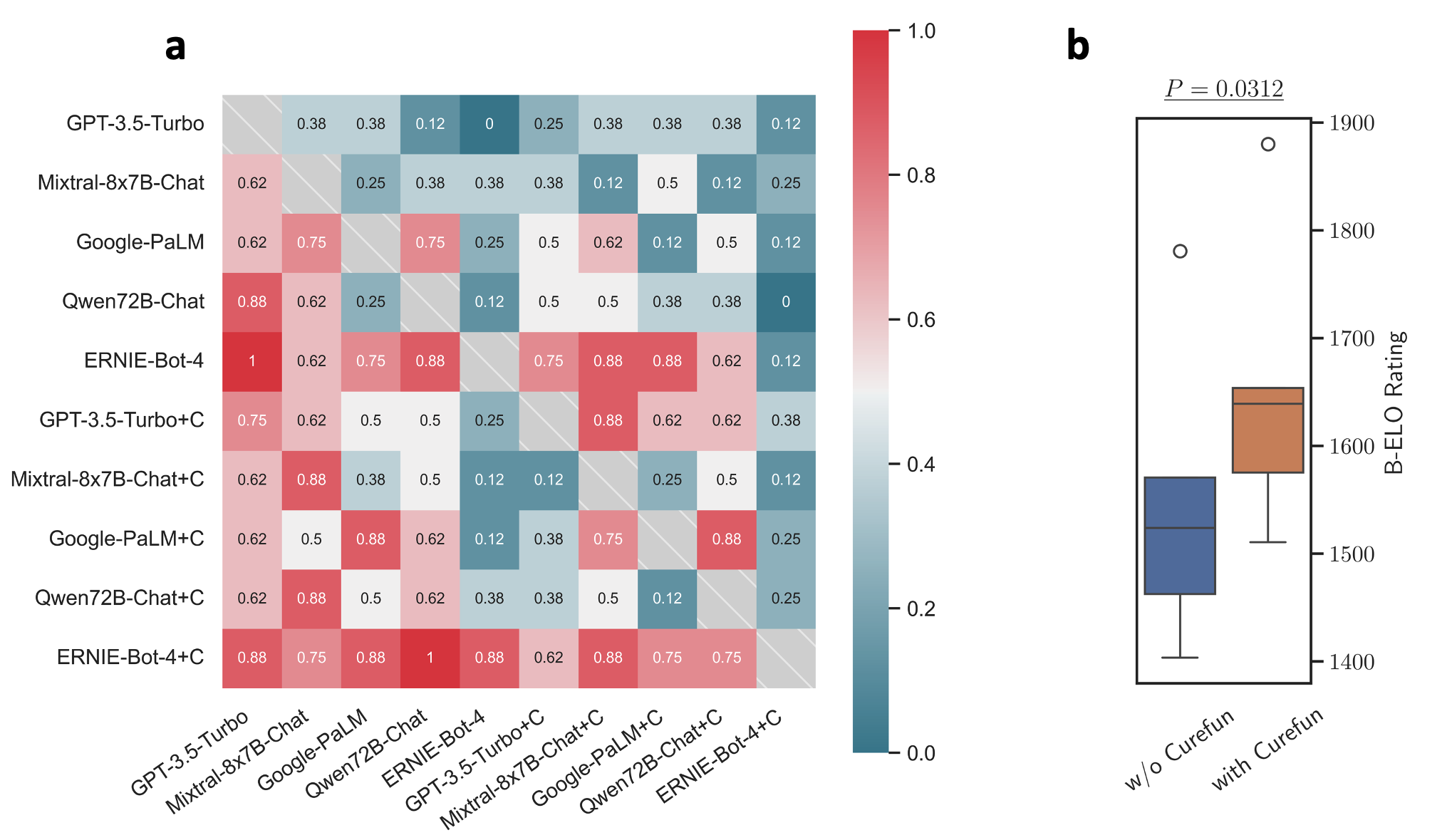}}
        \caption{(a) Heatmap of pairwise comparison for LLMs in acting SPs. The left-side labels represent the LLM used as ``player'', while the bottom labels represent the ``opponents''. ``+C'' denotes the corresponding model is collaborating with our framework. (b) The B-ELO score distribution with or without our framework, P < 0.05, one-sided Wilcoxon's rank-sum test.}
        \label{fig:SP-winrate}
    \end{figure}

    In order to further investigate the effectiveness of \textit{LLM as SP}, we conducted pairwise comparisons of various models following the setup of Chatbot Arena~\cite{zheng2023judging}.
    We employed GPT-4~\cite{achiam2023gpt}, which currently demonstrates the best performance, as the judge and selected the superior performer among each pair of SP dialogues.
    The results of the win rates are presented in Figure \ref{fig:SP-winrate}.
    Furthermore, we employed Bootstrap ELO rating (B-ELO) to calculate the stable capacity ratings of each model, considering both unordered and sufficient competition~\cite{duan2023botchat}.
    We computed vanilla ELO~\cite{elo1967proposed} ratings using the configuration of $initial rate = 1600$, $K = 100$, and then recomputed ELO scores 1,000 times in the randomly shuffled comparison order.
    The median of the 1000 ELO scores was adopted as the final score, as depicted in Table \ref{tab:elo}.

    From Figure~\ref{fig:SP-winrate} and Table \ref{tab:elo}, it can be observed that ERNIE-Bot-4 combined our framework achieved the best performance on the \textit{LLM as SP} task.
    This could be attributed to the fact that our case scripts are all in Chinese language, and ERNIE bot is one of the earliest LLMs specifically designed for Chinese corpora.
    Conversely, Mixtral is primarily built on datasets of Indo-European languages, hence the performance lags a bit.
    Disregarding the individual idiosyncrasies of these LLMs, it is noteworthy that our framework consistently and significantly improved performance across all backbone LLMs (Figure \ref{fig:SP-winrate}b; $P < 0.05$, one-sided Wilcoxon’s rank-sum test).
    Specifically, integrating our framework into GPT-3.5-Turbo resulted in a $250.18$-point increase in B-ELO score for the SP role-playing capacity, signifying a substantial advancement.

    \begin{table}[htbp]
        \centering
        \caption{B-ELO Ratings for LLM's capacity to play the role of SP.}
        \begin{tabular}{lcc}
            \toprule
            \multicolumn{1}{c}{\multirow{2}[0]{*}{\textbf{Model}}} &
            \multicolumn{2}{c}{\textbf{B-ELO}} \\
            \cmidrule(lr){2-3}
            & \multicolumn{1}{c}{w/o Ours} & \multicolumn{1}{c}{with Ours} \\
            \midrule
            Mixtral-8x7B  & 1462.40                      & 1510.60(\textbf{+48.20})      \\
            Qwen72B       & 1523.93                      & 1575.20 (\textbf{+51.27})     \\
            PaLM          & 1570.91                      & 1639.07 (\textbf{+68.16})     \\
            GPT-3.5-Turbo & 1403.54                      & 1653.72 (\textbf{+250.18})    \\
            ERNIE-Bot 4   & 1780.88                      & 1880.15 (\textbf{+99.27})     \\
            \bottomrule
        \end{tabular}%
        \label{tab:elo}%
    \end{table}

    Indeed, all of these LLMs have exhibited strong capabilities in role-playing.
    However, there have been some discrepancies in the role of SP. For example, GPT-3.5 demonstrates its high-level ethics and frequently flips the role as doctor following lengthy dialogues.
    Whereas our framework can consistently remind the model of the current task and the attributes it should consider.
    This feature is one of the key reasons why our framework has significantly enhanced the performance of GPT-3.5. Furthermore, the preferences of the judge, GPT-4, may introduce some biases.
    For instance, while the patient portrayed by PaLM is vivid and professional, GPT-4 often perceives it as lacking in detail due to PaLM's tendency to generate shorter responses.
    Overall, these comparative experiments consistently demonstrate the effectiveness of our framework in the scenario of \textit{LLM as SP}.

    \subsection{Evaluate Assessment}\label{sec:assessment}

    To validate the efficacy of automated assessment module within our proposed framework, we applied Curefun to evaluate the previously collected dialogue histories, consisting of a total of 80 records in 8 specific cases.
    Furthermore, an expert human evaluator was hired to independently fill out checklists and provide the grade for the aforementioned medicine dialogues, based on traditional scoring standard.
    We compared the scores obtained from both the automated scoring program and the human evaluators, sought to evaluate the consistency between these two assessment methods.
    To assess the degree of agreement between the human evaluators and our program, we employed two correlation measures: Spearman's rank correlation and Pearson correlation.
    These measures allow us to quantify the strength and direction of the relationship between the two sets of scores.

    \begin{table}[htbp]
        \centering
        \caption{Comparison of correlation measures between human evaluators' scores and program evaluators' scores.}
        \begin{tabular}{lcccc}
            \toprule
            \multicolumn{1}{c}{\multirow{2}[0]{*}{\textbf{Item}}} &
            \multicolumn{2}{c}{\textbf{Spearman's Rank}} & \multicolumn{2}{c}{\textbf{Pearson's}} \\
            \cmidrule(lr){2-3}
            \cmidrule(lr){4-5}
            & \multicolumn{1}{c}{Correlation} & \multicolumn{1}{c}{p-value} & \multicolumn{1}{c}{Correlation} & \multicolumn{1}{c}{p-value} \\
            \midrule
            Case1 & 0.954                           & 0.000                       & 0.927                           & 0.000                       \\
            Case2 & 0.655                           & 0.040                       & 0.943                           & 0.000                       \\
            Case3 & 0.822                           & 0.004                       & 0.903                           & 0.000                       \\
            Case4 & 0.810                           & 0.004                       & 0.820                           & 0.004                       \\
            Case5 & 0.803                           & 0.005                       & 0.765                           & 0.010                       \\
            Case6 & 0.758                           & 0.011                       & 0.772                           & 0.009                       \\
            Case7 & 0.832                           & 0.003                       & 0.809                           & 0.005                       \\
            Case8 & 0.820                           & 0.004                       & 0.832                           & 0.003                       \\
            \bottomrule
        \end{tabular}
        \label{tab:scoring}
    \end{table}

\begin{figure}[ht]
        \centering{\includegraphics[width=.7\linewidth]{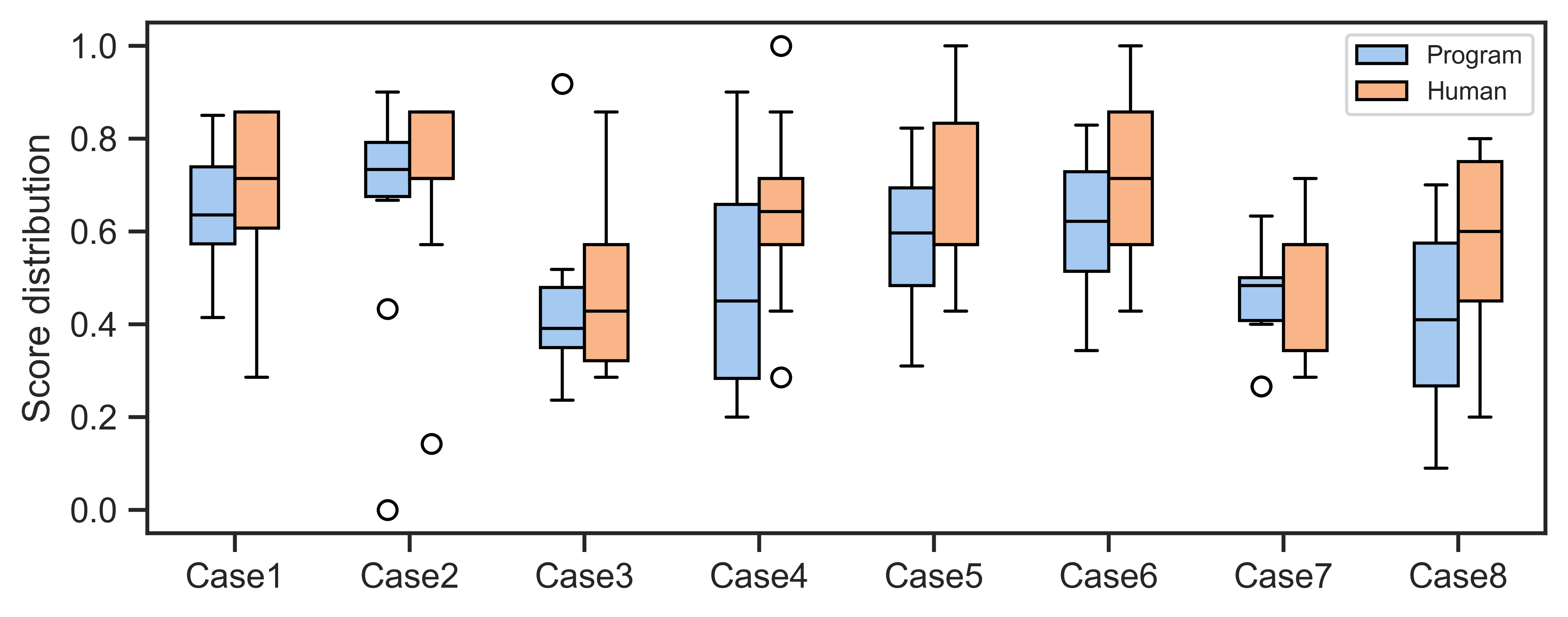}}
        \caption{The distribution of scores from program evaluator and human evaluator.}
        \label{fig:score-tester}
    \end{figure}

    Table~\ref{tab:scoring} shows Spearman's rank correlation and Pearson correlation between human evaluators' and our program's assessment scores.
    Overall, both Spearman's rank correlation and Pearson correlation coefficients are consistently closed to 1 (on average 0.81 and 0.85, respectively), suggesting a high degree of agreement between the two sets of scores.
    And the p-values associated with both correlation tests were found to be less than 0.05 for each group of cases.
    Those obtained correlation coefficients indicate a strong positive relationship between the human evaluators' scores and the scores generated by our program.
    The high correlations and statistically significant p-values suggest that our program's scores align closely with the judgments made by human evaluators.
    The score distribution depicted in Figure~\ref{fig:score-tester} further supports our analysis.
    The distribution appears to be symmetrical and concentrated around a central value, indicating a consistent and reliable scoring mechanism.
    This indicates that our automated assessment method produces reliable and accurate assessments, making it a suitable alternative to human evaluators in SP tests.

    \subsection{Evaluate LLMs as VDs}\label{sec:vds}

    Based on our comprehensive VSP, we have the capability to evaluate various public-available LLMs and test their standardized diagnostic interviewing abilities~\cite{REDDY2023101304}, which is a novel perspective due to its dynamic character.
    We integrate Curefun into the to-be-evaluated LLMs' chat flow as the role of users, and through multi-turn dialogues in predefined diagnosis scenarios, we stimulate LLMs to conduct medicine interviews with our VSP. Ultimately, the entire dialogue histories are scored using aforementioned assessment method, which records and analyzes various indicators during the conversation process.
    Except for the scores used to directly evaluate the VDs' diagnostic abilities, we also define several non-scoring indicators to provide more reference information for analyzing the VDs' performance.
    These indicators include:
    \begin{itemize}[left=0pt, itemsep=0pt, topsep=0pt]
        \item \textbf{Information Density}: Measures the density of entities and medical entities contained in the dialogue, indicating the professionalism and information density of the conversation.
        \item \textbf{Emotional Tendency}: Measures the emotional inclination in the dialogue, judging the emotional polarity of the conversation through an emotion analysis model, indicating the friendliness of the doctor role in the dialogue.
        \item \textbf{Response Length}: The length of the doctor's response in the dialogue, indicating the level of detail in the doctor's response.
        \item \textbf{Turn Number}: The total number of turns in the dialogue.
    \end{itemize}
    These indicators provide a comprehensive evaluation of the VDs' diagnostic abilities and help to analyze their performance in detail.

    We employ several kinds of LLMs as VDs, including SOTA business LLMs (ChatGPT and ERNIE-4-Bot), open-sourced general LLMs (Llama2-70B~\cite{touvron2023llama}, Mixtral-8x7B~\cite{jiang2024mixtral}), and medical-specific LLMs (BianQue-2 (6B)~\cite{chen2023bianque}, DISC-MedLLM (13B)~\cite{bao2023disc}), to evaluate their diagnostic abilities.
    Those LLMs are evaluated in the same 8 cases as the SPs, and each experiment is repeated 5 times to ensure the stability and reliability of the results.
    We also recruited human evaluators with different medical backgrounds to participate in the evaluation as a reference group.
    The human evaluators are classified into three categories: non-medical background folks, clinical medicine students, and experts, each of group contains 3 individuals.
    The non-medical background human evaluators have no medical knowledge, the clinical medicine students are second-year or third-year medical student, and the experts are senior physicians with rich clinical experience.
    The evaluation was conducted in a controlled environment, with each evaluator interacting with our SP chatbot.

    \begin{table}
        \centering
        \caption{Model Comparison}\label{tab:VDS}
        \resizebox{\linewidth}{!}{
        \begin{tabular}{lccccc}
            \toprule
            \multicolumn{1}{l}{\multirow{2}[0]{*}{\textbf{Model}}} & \multicolumn{1}{p{2cm}}{\centering \textbf{Information \\Density}}                        & \multicolumn{1}{p{2cm}}{\centering \textbf{Emotional \\Tendency}}                      & \multicolumn{1}{p{2cm}}{\centering \textbf{Response Length}} & \multicolumn{1}{p{2cm}}{\centering \textbf{Turn \\Number}}                    & \multicolumn{1}{p{2cm}}{\centering \textbf{Overall \\Score}} \\ \midrule
            Llama2-70B                 & 0.02 & 0.93 & 2996.17  & 5.39  & 0.34 \\
            Mixtral-8x7B                       & 0.11 & 0.69 & 647.50  & 5.88  & 0.33 \\
            BianQue-2 (6B)               & 0.14 & 0.69 & 178.00  & 4.25  & 0.25 \\
            DISC-MedLLM (13B)    & 0.15 & 0.66 & 361.87  & 4.57  & 0.43 \\
            ERNIE-4-Bot & 0.13 & 0.69 & 662.09  & 5.70 & 0.37 \\
            ChatGPT (3.5-turbo)                    & 0.15 & 0.70 & 371.74  & 7.74 & 0.51 \\ \midrule
            Human (Non-medical background)                    & 0.15 & 0.71 & 125.00  & 8.45 & 0.45 \\
            Human (Clinical medicine student)                    & 0.19 & 0.48 & 212.99  & 23.11 & 0.72 \\
            Human (Expert)                    & 0.27 & 0.56 & 135.75  & 13.38 & 0.78 \\
            \bottomrule
        \end{tabular}
        }
    \end{table}

    Table~\ref{tab:VDS} presents the evaluation results of the performance and metrics of LLMs acting as VDs. The results show that ChatGPT obtains the highest score among all LLMs.
    DISC-MedLLM achieves the second-highest score, indicating that medical-specific LLMs could exhibit better diagnostic abilities than general LLMs. However, BianQue-2 achieves a relative lower score, which may attribute to its smallest parameter scale thus lack of multi-turn dialogue ability.
    Besides, human evaluators, especially the experts, outperform all LLMs in terms of diagnostic ability, indicating that LLMs still have room for improvement in simulating real-world medical scenarios.

    From the non-scoring indicators, we can observe the diversified preferences of LLMs and human-beings.
    For example, LLMs, especially Llama2-70B, tend to generate long responses while human evaluators like to talk in relative shorter length.
    All LLMs tend to end the conversation in fewer rounds, while human evaluators prefer more rounds of conversation, especially clinical medicine students have an average of 23 rounds of conversation.
    This may be because human evaluators tend to communicate with patients to obtain information through in-depth conversation, while LLMs prefer to solve problems in fewer rounds to reduce costs.
    Additionally, it can be noted that LLMs consistently perform well in emotional tendency, or the degree of friendliness in communication with patients.
    This is because LLMs are usually given a certain degree of friendliness during alignment training.
    The friendliness of human evaluators in this indicator varies, which may be due to the fact that human evaluators are influenced by more factors during the conversation, such as empathy for patients and their individual career experience\footnote{An interesting finding is that human doctors tend to shift towards negative sentiment polarity over time during the entire test, whereas the LLMs remain positive throughout. This suggests the potential of LLMs for greater stability than humans in medical consultation.}.

    In conclusion, although LLM falls short of human experts in our SP testing, it has already reached a level comparable to ordinary individuals without medical backgrounds.
    Additionally, LLM consistently maintains a superior emotional indicator when interacting with patients compared to humans.
    This result indicates that LLMs can assist human doctors to some extent in medical consultations, highlighting the potential of LLMs as pre-diagnostic and triage tools.

    It's worth noting that SPs and VDs, considered complementary tasks, do not align perfectly in real-world scenarios, as confirmed by our diagnostic evaluation of various LLMs. Highly performing LLMs in diagnostic QA benchmarks do not meet expectations in our VSP setting.
    Firstly, SPs are primarily designed for education and rely on standardized examinations, prioritize comprehensive and standardized communication, while real patient interactions prioritize efficiency and flexibility.
    This disparity exists not only between VSPs and VDs but also in real-world clinical education and practice.
    Therefore, aspiring students are advised to gain practical experience to align with real-world medical scenarios.
    Based on these observations, we can develop an LLM training process integrating VSP and VD tasks to facilitate self-improvement, as explored in a recent study~\cite{tu2024towards}.

    \section{Method}

    In this section, we describe the detailed methodology and main components of our proposed framework.
    We developed Curefun from the ground up, including the data processing pipeline, the graph-driven context-adaptive SP chatbot, the LLM-based automatic assessment methods, and some infrastructure for supporting our proposed framework and improving the user experience.
    The overview of our framework is illustrated in Figure~\ref{fig:overall}.

    \subsection{Data Processing}

\begin{figure}[ht]
        \centering{\includegraphics[width=.8\linewidth]{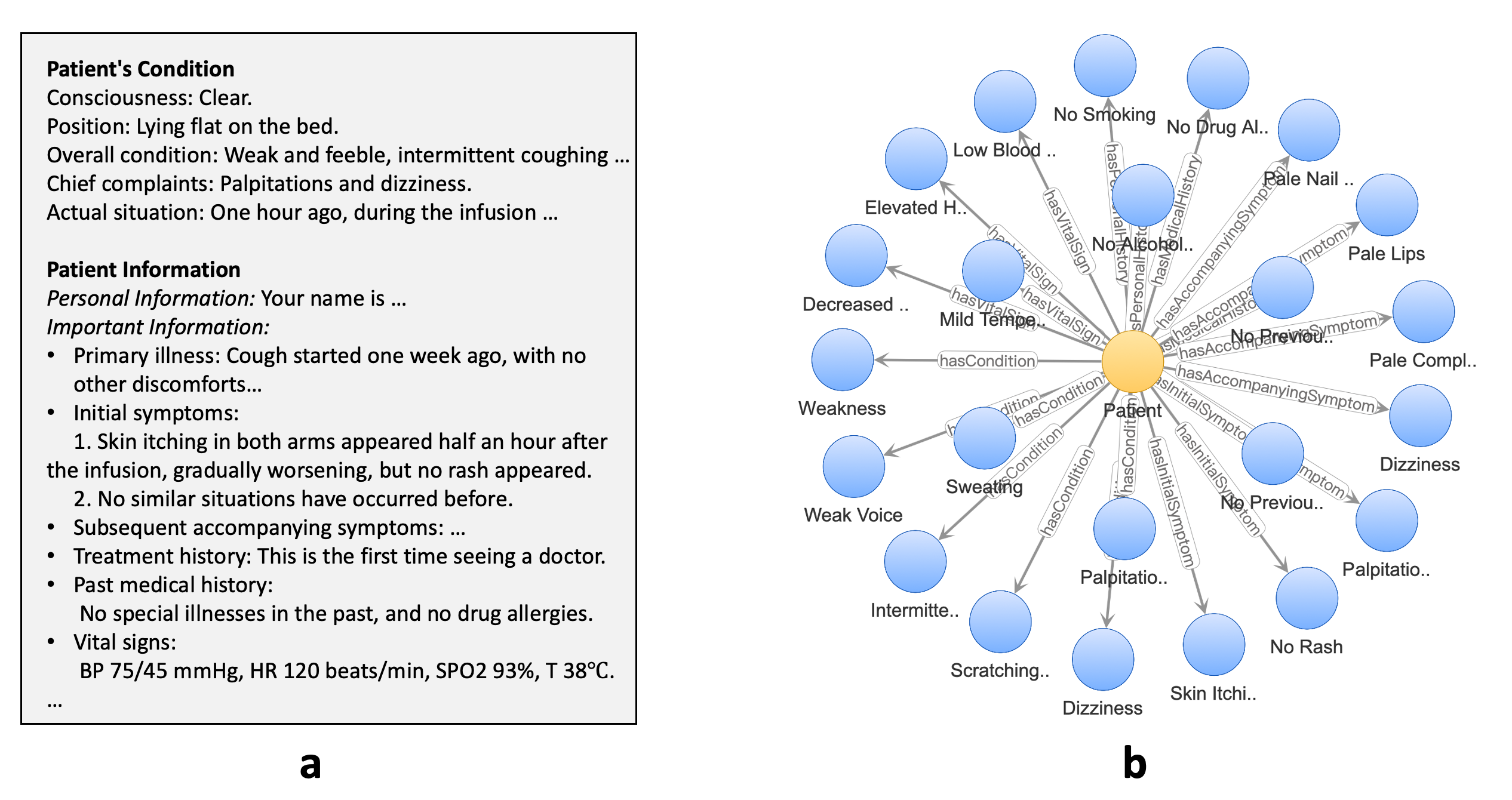}}
        \caption{Preview of a SP case. (a) Original case script/template. (b) Extracted case graph.}
        \label{fig:case-example}
    \end{figure}

    For each SP case, we utilize Named Entity Recognition (NER)~\cite{li2020survey} and relation extraction~\cite{wang2021relation} models to extract entities and relationships between entities from the SP script, forming the skeleton graph of the corresponding case.
    Then, we employ open information extraction method~\cite{wang2018clinical, li2022gbuilder} to extract the attributes and the corresponding attribute values associated with entities in the skeleton graph, creating a case graph, which can describe the internal relationships and various attributes of the case.
    Specifically, we trained a BERT-based NER model~\cite{rasmy2021med, devlin-etal-2019-bert} on both general corpus and medical datasets.
    This model is used to extract entities in the general domain and the medical domain~\cite{bose2021survey}, including medications, symptoms, and diseases, etc.
    Then we employed a simple BERT-based relation classification model~\cite{roy2021incorporating} to detect relationships between various entities~\cite{hahn2020medical}.
    An open attribute extraction model~\cite{li2023attgen} was introduced to supplement the attributes of medical entities.
    Some literal values that cannot be contained in the entity-relation skeleton, such as body temperature and blood pressure, will be extracted as attributes and linked to the skeleton graph.
    Figure~\ref{fig:case-example} presents a constructed case graph through aforementioned processes.

    The purpose of using aforementioned information extraction methods to construct a case graph is to utilize the RAG manner.
    When a student mentions relevant symptoms or relationships, the RAG method stimulate model to retrieve relevant subgraphs of entities, relationships, and attributes from the whole case graph.
    This effectively reduces the length of the case scripts occupying the LLM input buffer, alleviating the degradation of generation quality caused by excessively long form dialogue.
    It enhances the quality and coherence of dialogue generation in VSP's conversation~\cite{wei2018build, xu2023recomp}.

    Except constructing the case graph, we also preprocess the corresponding SP checklist for automatic assessment, which is detailed in Section~\ref{sec:auto-assessment}.

    \subsection{Graph-Driven Context-Adaptive SP Chatbot}

\begin{figure}[ht]
        \centering{\includegraphics[width=.95\linewidth]{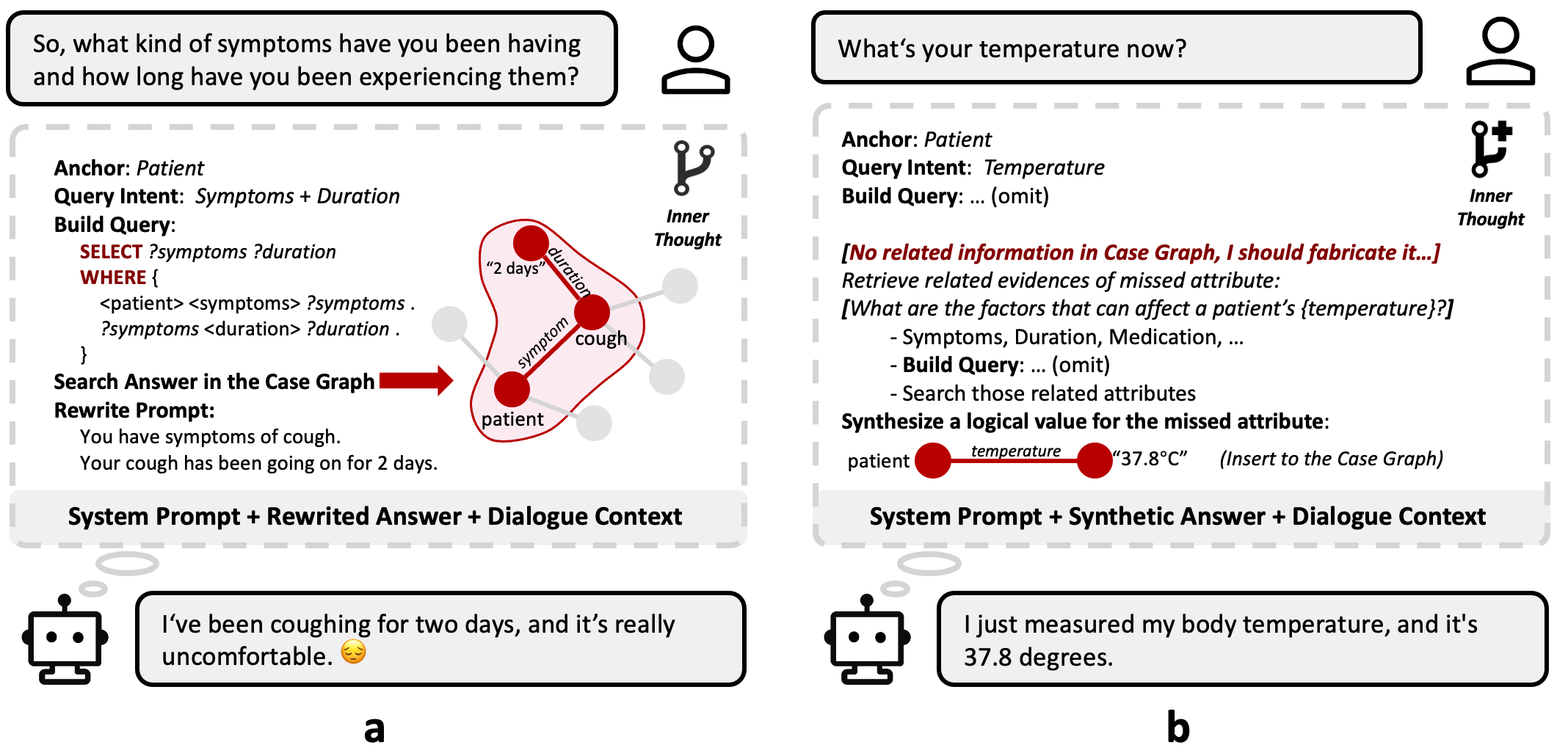}}
        \caption{Overview of Curefun's context-adaptive mechanism. (a) A Q\&A sample in dialogue flow demonstrates how Curefun achieves precise replies to user queries through the ``ERRG'' process. (b) An example of a single-turn conversation illustrating how Curefun synthesizes coherent responses and maintains dialogue consistency even when confronted with missing attributes.}
        \label{fig:context-dynamic}
    \end{figure}

    To enhance the dialogue quality and coherence of the VSP, we propose the integration of a graph-driven context-adaptive mechanism into the LLM-based chatbot.
    This mechanism aims to dynamically adjust the flow of the dialogue by utilizing a controlled and structured graph memory.
    The process is illustrated in Figure~\ref{fig:context-dynamic} (a), which consists of four main steps: \textbf{E}xtract, \textbf{R}etrieve, \textbf{R}ewrite, and \textbf{G}enerate (ERRG).
    In the ``Extract'' step, the chatbot extract the core entities and possible relations from the user's input and the ongoing dialogue context.
    These extracted mentions are then linked to the corresponding entity nodes in the case graph through substring matching and string proximity.
    In the ``Retrieve'' step, the chatbot utilizes the LLM to generate a formal query based on the extracted entities and relations, then executes the query on the case graph to retrieve the relevant subgraph.
    In the ``Rewrite'' step, the chatbot transforms the retrieved subgraph into natural language form, thereby providing context-aware evidence for generating the response.
    Finally, in the ``Generate'' step, the chatbot generates an optimal response by considering the retrieved evidence, the current dialogue context, and the user's input.

    Based on ``ERRG'' process, the chatbot can answer user queries accurately and coherently based on case graph, which represents the known information from the SP case.
    However, in the setting of open-world dialogue, SP case scripts cannot contain all the information that the user may ask.
    In this situation, the chatbot needs to fabricate rational attributes and entities to maintain the dialogue coherence.
    For preventing the inconsistency response about fictional information, and to prevent the potential conflict between the generated information and the known information in the case graph, which may lead to hallucinations, we introduce a controlled fictional information generation approach.
    As shown in Figure~\ref{fig:context-dynamic} (b), the chatbot can generate a coherent response even when the user asks about missing attributes.
    The chatbot can synthesize rational attributes based on the known information and the LLM's internal common sense knowledge, and then persist the generated information in the case graph to maintain dialogue consistency.
    Finally, the chatbot can provide a coherent and informative response to the user's query, even when the information is not explicitly provided in the case graph.

    It is worth noting that our proposed framework, Curefun, serves as a model-agnostic shell capable of accommodating a wide range of chat-oriented LLMs, as demonstrated in Table~\ref{tab:elo}.
    However, in light of considerations pertaining to usability, cost, and efficacy, we employ a Llama-13B based checkpoint~\cite{fengshenbang} as the backbone model in the production.
    Subsequently, the automatic assessment and the corresponding experiments in Section~\ref{sec:assessment} and Section~\ref{sec:vds} were conducted using this model as the foundation.

    \subsection{LLM-based Automatic Assessment}\label{sec:auto-assessment}

\begin{figure}[ht]
        \centering{\includegraphics[width=.85\linewidth]{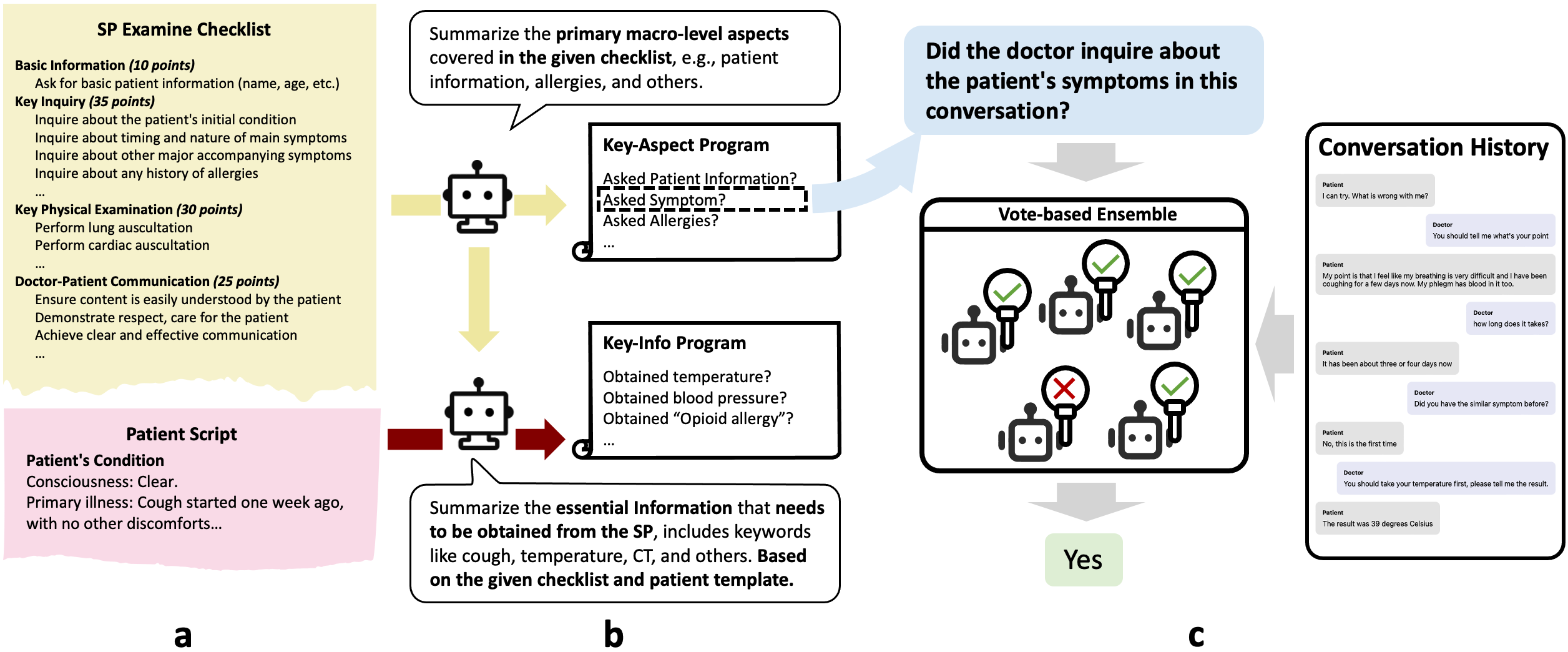}}
        \caption{A sample of an assessment. (a) Original case checklist. (b) Processed automatic scoring program. (c) Vote-based LLMs ensemble.}
        \label{fig:assessment-sample}
    \end{figure}

    In order to comprehensively and dynamically evaluate students' performances and expand to large-scale assessments, we utilize LLM to analyze, extract, and summarize the various rating items in the original checklist, and eventually forming two elements for scoring:
    \begin{itemize}[left=0pt]
        \item \emph{Aspects}: These aspect items necessitate proactive inquiry by the student in order to accrue scoring credit. E.g., whether the student asked about the patient's genetic medical history.
        \item \emph{Information}: These key information items must be elicited from patient through inquiry to obtain scoring credit. E.g., specific symptoms and important physiological indicators.
    \end{itemize}
    The scoring elements at granularity hierarchical level simplify the complex checklist into single tasks compatible with LLMs, facilitating more accurate assessment.
    An example of a structured assessment checklist is illustrated in Figure~\ref{fig:assessment-sample}.
    This approach allows us to seamlessly transform the complex checklist into an automated evaluation procedure.

    After generating the automatic scoring program, we employed a series of LLMs to perform scoring for each item and ensemble their opinion through voting, as shown in Figure~\ref{fig:assessment-sample}(c).
    After voting for all items in the program, we can obtain a reliable and stable score for the clinical history.
    In practice, we used 5 different Llama-13B checkpoints for ensemble. The proportions of \emph{Aspect} and \emph{Information} in the total score are $0.3$ and $0.7$, respectively, according to the checklist in the actual SP examination.
    The scores obtained through this approach allow for a comprehensive assessment of students' mastery of various aspects of the SP case in the examination.
    Furthermore, it effectively prevents attacks targeting specific exam items, as obtaining a valid score for ``Information'' requires eliciting it through inquiry and cannot be achieved by exhaustively listing it in endless questions.
    In addition to assessing the mastery of ``Aspects'' and ``Key Information'', we have also defined several metrics, including the emotional inclination of the dialogue, the entity-related information density, the dialogue length, and the number of dialogue turns.
    These metrics, as shown in Table~\ref{tab:VDS} in Section~\ref{sec:vds}, are difficult to quantify and may reduce the robustness of assessment.
    Therefore, they serve as non-scoring indicators to provide reference information for educators and students.
    The experimental results presented in Section~\ref{sec:assessment} demonstrate that our proposed assessment programs yield scores consistent with human grading.

    \subsection{Auxiliary Modules}

    We introduce several auxiliary modules that enhance the performance of our proposed framework.

    \textbf{TTS/STT Models}.
    To facilitate natural conversations between students and our proposed system, we integrate a Text-to-Speech (TTS)~\cite{tan2021survey} and Speech-to-Text (STT)~\cite{malik2021automatic} module into our framework.
    The TTS module converts the generated patients' responses into realistic speech, providing a more immersive experience for students.
    Conversely, the STT module transcribes the students' spoken inquiries into text format, enabling the LLMs to process and respond accordingly.
    By incorporating TTS and STT technologies, our framework bridges the gap between written dialogue and spoken interaction, enhancing the authenticity and effectiveness of the VSP experience.

    \textbf{Graph Database}.
    To enable efficient storage and retrieval of character settings and LLM-fabled attributes, we employ the graph database~\cite{zou2011gstore} as a knowledge repository in our framework.
    The graph database can efficiently organize medical concepts, conditions, and symptoms as nodes, while the relationships between them are represented as edges~\cite{angles2008survey}.
    Hereby we introduce the standardized Resource Description Framework (RDF)~\footnote{\url{http://www.w3.org/TR/rdf-sparql-query/}} format to represent the structured patient information, then use SPARQL~\cite{DBLP:conf/semweb/LiZH022} queries to retrieve the relevant information in graph database.
    This structure allows for flexible querying and navigation of patient information, facilitating the generation of accurate and contextually relevant responses by the LLMs.

    \textbf{LLM Server}.
    To support the computational requirements of our framework, we introduce the dedicated and high-performance LLM server~\cite{kwon2023efficient}.
    The server hosts the LLMs, adopting acceleration technique like page attention~\cite{kwon2023efficient} and speculative decoding~\cite{chen2023accelerating}, enabling efficient and parallel processing of student inquiries and SPs' responses.
    The LLM server plays a critical role in ensuring the overall performance and scalability of our framework.

%

    \bibliographystyle{unsrtnat}
    \bibliography{references}

\end{document}